\documentclass[a4paper,UKenglish,cleveref, autoref, thm-restate]{lipics-v2021}

\hideLIPIcs  

\graphicspath{{./images/}}

\title{Learning Size Independent Constraint Models from Input Data and Positive and Negative Example Solutions} 

\titlerunning{Learning Size Independent Constraint Models} 

\author{Helmut Simonis}{School of Computer Science and Information Technology, University College Cork, Ireland}{h.simonis@ucc.ie}{https://orcid.org/0009-0003-6117-2463}{}

\authorrunning{H. Simonis} 

\Copyright{Helmut Simonis} 



\category{} 

\relatedversion{} 

\nolinenumbers 

\usepackage{booktabs}
\usepackage{xcolor}
\usepackage{tikz}
\usetikzlibrary{calc}
\usetikzlibrary{positioning}
\usetikzlibrary{shapes}
\usetikzlibrary{shadows}
\usetikzlibrary{arrows.meta}
\usepackage{mdframed}
\usepackage{listings}
\usepackage{colortbl}

\definecolor{eclipseStrings}{RGB}{42,0.0,255}
\definecolor{eclipseKeywords}{RGB}{127,0,85}
\colorlet{numb}{magenta!60!black}

\usepackage{listings}
\usepackage{courier} 

\lstdefinelanguage{Mzn}
{
  morekeywords={
  array, par, var, opt, constraint, solve, satisfy, minimize,
  maximize, output, include, let, in, set, of, if, then, else, elseif, endif,
  ann, annotation, bool, enum, float, int, string, where, function,
  predicate, true, false, not, assert, trace,
  any, list, op, record, test, tuple, type
  },
  keywords=[2]{
  forall, exists, xor, xorall, iffall, clause,
  all_different, all_different_int, alldifferent, alldifferent_int,
  alldifferent_except_0, alldifferent_except, all_equal,
  nvalue, diffn,
  at_least, at_most, exactly, 
  count, count_eq, count_leq, count_geq, count_gt,
  global_cardinality, global_cardinality_closed,
  global_cardinality_low_up, global_cardinality_low_up_closed,
  element, regular, regular_nfa, table, inverse,
  bin_packing, bin_packing_capa, bin_packing_load, knapsack,
  cumulative, disjunctive, circuit, subcircuit, dpath,
  decreasing, increasing,
  strictly_decreasing, strictly_increasing, 
  lex_less, lex_lesseq, lex_greater, lex_greatereq, lex2, strict_lex2,
  value_precede, value_precede_chain,
  symmetry_breaking_constraint, implied_constraint, redundant_constraint,
  sort, arg_sort, among, sliding_sum,
  int_ne, int_lt_reif, int_lin_eq, int_lin_eq_reif, bool_lin_eq,
  bool_imply, array_bool_or,
  in_set, subset, superset, partition_set, member,
  abort,
  acosh, asin, atan, cos, cosh, sin, sinh, tan, tanh,
  array_intersect, array_union,
  array1d, array2d, array3d, array4d, array5d, array6d,
  bool2int, int2float, set2array,
  abs, div, mod, pow, exp, sqrt, ln, log, log2, log10,
  ceil, floor, round,
  min, max, length, product, sum,
  dom, dom_array, dom_size, fix, is_fixed,
  index_set, index_set_1of2, index_set_2of2,
  index_set_1of3, index_set_2of3, index_set_3of3, index_set_6of6,
  concat, reverse, join,
  lb, lb_array, ub, ub_array,
  show, show_int, show_float,
  card, intersect, union, diff, symdiff,
  is_defined_var, output_var, var_is_introduced, defines_var,
  promise_total, bounds, domain, bool_search, int_search, seq_search,
  set_search, input_order, first_fail, anti_first_fail, smallest,
  largest, occurrence, most_constrained, max_regret, indomain_min,
  indomain_max, indomain_middle, indomain_median, indomain,
  indomain_random, indomain_split, indomain_reverse_split,
  indomain_interval, outdomain_max, outdomain_median, outdomain_min,
  outdomain_random, complete
  },
  sensitive=true,
  basicstyle=\scriptsize,
  backgroundcolor=\color{white}, 
  commentstyle=\color[rgb]{0.9,0.1,0.1},
  keywordstyle=\color[rgb]{0,0.5,0},
  keywordstyle=[2]\color{blue},
  stringstyle=\color{orange},
  tabsize=2,
  frame=none,
  numbers=left,
  stepnumber=1,
  numberstyle=\tiny,
  numbersep=5pt,
  xleftmargin=0pt, 
  columns=fixed, 
  morecomment=[l]{\%},
  morestring=[b]",
  showstringspaces=false,
  mathescape=true,
  breaklines=true,
  breakatwhitespace=true,
  breakindent=10pt, 
  moredelim=**[is][\color{Melon}]{@}{@},
  escapeinside={{<@}{@>}}
}
\lstnewenvironment{mzn}[1][]{\lstset{language=Mzn,#1}}{}
\lstnewenvironment{mznno}[1][]{\lstset{language=Mzn,numbers=none,xleftmargin=0pt,#1}}{}

\lstdefinelanguage{json}{
    basicstyle=\normalfont\ttfamily,
    commentstyle=\color{eclipseStrings}, 
    stringstyle=\color{eclipseKeywords}, 
    numbers=left,
    numberstyle=\tiny,
    stepnumber=1,
    numbersep=8pt,
    showstringspaces=false,
    breaklines=true,
    frame=lines,
    backgroundcolor=\color{white}, 
    string=[s]{"}{"},
    comment=[l]{:\ "},
    morecomment=[l]{:"},
    literate=
        *{0}{{{\color{numb}0}}}{1}
         {1}{{{\color{numb}1}}}{1}
         {2}{{{\color{numb}2}}}{1}
         {3}{{{\color{numb}3}}}{1}
         {4}{{{\color{numb}4}}}{1}
         {5}{{{\color{numb}5}}}{1}
         {6}{{{\color{numb}6}}}{1}
         {7}{{{\color{numb}7}}}{1}
         {8}{{{\color{numb}8}}}{1}
         {9}{{{\color{numb}9}}}{1}
}
\definecolor{pantone174-6}{RGB}{127,138,154}
\definecolor{insight-gray}{RGB}{127,138,154}
\definecolor{pantone117-8}{RGB}{0,107,148}
\definecolor{insight-blue}{RGB}{0,107,148}
\definecolor{pantone127-4}{RGB}{114,202,195}
\definecolor{insight-aqua}{RGB}{114,202,195}
\definecolor{pantone157-8}{RGB}{146,200,62}
\definecolor{insight-lime}{RGB}{146,200,62}
\definecolor{pantone126-8}{RGB}{0,105,106}
\definecolor{insight-forestgreen}{RGB}{0,105,106}
\definecolor{pantone151-8}{RGB}{76,183,72}
\definecolor{insight-green}{RGB}{76,183,72}
\definecolor{pantone119-1}{RGB}{176,215,227}
\definecolor{insight-babyblue}{RGB}{176,215,227}

\definecolor{pantone10-7}{RGB}{255,198,52}
\definecolor{insight-yellow}{RGB}{255,198,52}
\definecolor{pantone24-8}{RGB}{246,135,31}
\definecolor{insight-orange}{RGB}{246,135,31}
\definecolor{pantone46-8}{RGB}{193,49,34}
\definecolor{insight-burntorange}{RGB}{193,49,34}
\definecolor{pantone74-7}{RGB}{186,52,111}
\definecolor{insight-pink}{RGB}{186,52,111}
\definecolor{pantone92-7}{RGB}{105,44,122}
\definecolor{insight-plum}{RGB}{105,44,122}
\definecolor{pantone72-16}{RGB}{145,0,69}
\definecolor{insight-maroon}{RGB}{145,0,69}
\definecolor{pantone120-2}{RGB}{128,184,200}
\definecolor{insight-dodgerblue}{RGB}{128,184,200}
\definecolor{pantone149-4}{RGB}{115,175,117}
\definecolor{insight-palegreen}{RGB}{115,175,117}
\definecolor{pantone124-16}{RGB}{0,152,153}
\definecolor{insight-turquoise}{RGB}{0,152,153}
\definecolor{pantone116-14}{RGB}{0,149,191}
\definecolor{insight-royalblue}{RGB}{0,149,191}
\definecolor{pantone102-7}{RGB}{61,91,169}
\definecolor{insight-purple}{RGB}{61,91,169}
\definecolor{pantone108-16}{RGB}{0,52,98}
\definecolor{insight-darknavy}{RGB}{0,52,98}

\EventEditors{John Q. Open and Joan R. Access}
\EventNoEds{2}
\EventLongTitle{42nd Conference on Very Important Topics (CVIT 2016)}
\EventShortTitle{CVIT 2016}
\EventAcronym{CVIT}
\EventYear{2016}
\EventDate{December 24--27, 2016}
\EventLocation{Little Whinging, United Kingdom}
\EventLogo{}
\SeriesVolume{42}
\ArticleNo{23}

\begin{document}

\maketitle

\begin{abstract}
In this paper we describe a Constraint Acquisition Tool that generates size independent constraint programs from input data and positive and negative example solutions of multiple problem instances. The system finds size specific models based on typical constraint patterns over global and binary constraints, then generalizes the models to a size independent model that can be run on unseen input data. The system produces MiniZinc code as well as a textual description of the model. We evaluate the tool on the example problems of the PTHG21 Challenge and show the feasibility of the approach.
\end{abstract}

\section{Introduction}
\label{sec:introduction}

Constraint Acquisition~\cite{DBLP:journals/ai/BessiereKLO17} is the process of defining a constraint model from a set of example solutions and non-solutions. In most of the existing literature, a model of a single problem instance is learned, which can then only be used to generate new solutions of the same problem, or perhaps the same problem with different domain bounds. In most practical CP applications, the actual constraint model heavily depends on input data. Consider a scheduling problem for a factory. Every day the set of orders that need to be scheduled will be different, and from time to time new products or processes will be introduced. A CP model for this factory needs to be able to create the appropriate constraint model for a given dataset, and change the specific constraints of the instance from one day to the next, due to different process steps or resources used. If we want to acquire a generic constraint model in this scenario, we have to understand how the number of variables, their domains, and the constraints depend on the input data. Having acquired the model, we then should be able to apply the model to a new, previously unseen set of input data, and solve the resulting model. This is a more challenging problem than the use case previously discussed in constraint acquisition, but, as this paper shows, this task is not out of reach at this point. 

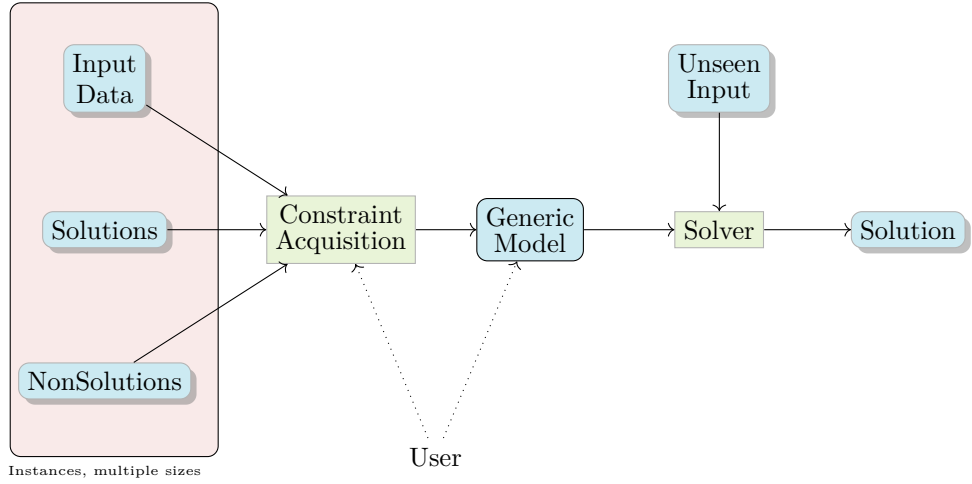
\begin{figure}[htbp]
\caption{\label{fig:intendedusecase}Intended Use Case}
\begin{tikzpicture}[xscale=2.5,yscale=2,
inputdata/.style={fill=insight-burntorange!10,draw=black,rounded corners},
program/.style={fill=insight-lime!20,draw=black!30},
documents/.style={fill=insight-royalblue!20,draw=black!30,rounded corners,drop shadow},
document/.style={fill=insight-royalblue!20,draw=black,rounded corners}]
\draw[inputdata] (0.25,1.5) rectangle (1.35,4.5);
\node[documents] (input) at (0.75,4) {\shortstack{Input\\Data}};
\node[documents] (solutions) at (0.75,3) {Solutions};
\node[documents] (nonsolutions) at (0.75,2) {NonSolutions};
\node[below] at (0.75,1.5) {\tiny Instances, multiple sizes};
\node[program] (ca) at (2,3) {\shortstack{Constraint\\Acquisition}};
\node[document] (model) at (3,3) {\shortstack{Generic\\Model}};
\node[documents] (unseen) at (4,4) {\shortstack{Unseen\\Input}};
\node[program] (solver) at (4,3) {Solver};
\node[documents] (solution) at (5,3) {Solution};
\node[] (user) at (2.5,1.5) {User};
\draw[->] (input) -- (ca);
\draw[->] (solutions) -- (ca);
\draw[->] (nonsolutions) -- (ca);
\draw[->] (ca) -- (model);
\draw[->] (model) -- (solver);
\draw[->] (unseen) -- (solver);
\draw[->] (solver) -- (solution);
\draw[->,dotted] (user) -- (ca);
\draw[->,dotted] (user) -- (model);
\end{tikzpicture}
\end{figure}

The general use case we want to model is shown in Figure~\ref{fig:intendedusecase}. We are given a set of input data, solutions and non-solutions for multiple problem instances of different sizes. Our constraint acquisition tool produces a generic model based on those problem instances. This model can then be fed, together with new, unseen input data, into an existing constraint solver to produce a solution for this new instance. The user can interact with the constraint acquisition to accept or reject certain constraints, in addition the users should be able to understand and extend the generated model if they so wish.

This leads to the following set of properties that a generic constraint acquisition tool should have: 

\begin{description}
\item[diverse structure] We must be able to handle not just a single set of unstructured variables, but a collection of variables, vectors and matrices, and possibly more complex structures. We can exploit this structural information to find likely constraints linking multiple structures, which would be difficult to identify in an unstructured list.
\item[hidden variables] The variables described in the solution may not describe all variables used in the constraint model. Some derived variables, especially cost variables, may be missing from the samples, with only aggregate cost values presented. Our system should reconstruct such missing, hidden variables in order to find the complete constraint model.
\item[generic] The model should not be specific to one given instance only, but apply to a series of instances which are specified by different input data and parameter values.
\item[consistent] The generated model should be consistent, so that all given positive samples are accepted, and all negative samples are rejected. We are currently not handling the case where solutions or non-solutions are labelled incorrectly.
\item[minimal] The model should not contain any constraints that are not required, in particular any subsumed constraint should be filtered out.
\item[transferable] We must be able to use the generated model on new, previously unseen input data, as long as the structure of the problem remains the same.
\item[executable] It is not enough to generate an abstract model, we should be able to execute the model with some existing constraint solver. In our case, we produce MiniZinc\cite{DBLP:conf/cp/NethercoteSBBDT07} code, which can then be run with a variety of back-end solvers.
\item[explainable] We should be able to explain the generated model to a user not familiar with specific constraint programming systems and languages. A natural language description of the model would provide a good initial explanation.
\item[efficient] As we are solving combinatorially hard problems, we will not be able to solve every new problem instance. We consider the generated model efficient if its runtime is comparable to a handwritten program for the same problem written in a high-level modelling language like MiniZinc.
\end{description}

The main contributions of this paper are: 
\begin{itemize}
\item introduction of a more realistic constraint acquisition use case driven by practical needs
\item generating size independent constraint models dealing with structured, positive and negative samples and input data
\item simplifying the concept of partition generator in ModelSeeker~\cite{DBLP:conf/cp/BeldiceanuS12} to allow for trivial generalization with an acceptable loss of flexibility
\item producing executable constraint programs in MiniZinc, whose performance is comparable to hand-crafted code for the same problem
\item producing a natural language, textual description of the generated model to explain the model to a user
\item demonstrating the feasibility of the approach on all benchmark problems defined for the PTHG21 Challenge~\cite{simonis_helmut_2021_5155465}
\end{itemize}  

The paper is structured in the following way. In the next Section we discuss related work, and influences on the current paper. In Section~\ref{sec:running example} we introduce a running example that we use throughout the presentation of the different elements of our system. Section~\ref{sec:system description} is the main part of paper, describing the different components of the tool. This is followed in Section~\ref{sec:evaluation} by an evaluation of the tool based on the problems in the PTHG21 Challenge. We conclude in Section~\ref{sec:conclusion} with a discussion of existing gaps, and future work.

\section{Related Work}
\label{sec:related work}

Work on constraint acquisition started in the early 2000s, with~\cite{DBLP:conf/aisc/Freuder02,DBLP:conf/cp/ColettaBOFOQ03,DBLP:conf/cp/BessiereCFO04}. Two main approaches developed, one (called "passive") based on labelled example solutions and non-solutions, and one (called "active") based on asking queries to a hypothetical user who knows the problem, but cannot express the constraints directly. A general overview of the literature is given in~\cite{DBLP:journals/ai/BessiereKLO17,DBLP:series/lncs/BessiereDHKLMNQW16}. The work on active constraint acquisition~\cite{DBLP:conf/ijcai/BessiereCOP07,DBLP:journals/corr/abs-2003-06649,DBLP:conf/cp/Tsouros0B19,DBLP:conf/cpaior/AddiBEL18,DBLP:conf/ijcai/ArcangioliBL16,DBLP:conf/ecai/BessiereCDLMB14,DBLP:conf/ijcai/BessiereCHKLNQW13,DBLP:journals/constraints/TsourosS20} centres on the question of how many queries need to be asked, and which form these queries can take. 

More recent work has looked at extending constraint acquisition to other problem domains, like Integer Programming~\cite{DBLP:series/lncs/RaedtDGB16,DBLP:conf/ictai/KumarTCR19,DBLP:conf/ijcai/KumarTR19,DBLP:journals/eor/PawlakL21,DBLP:journals/tec/PawlakK19,DBLP:journals/eor/PawlakK17}, SMT~\cite{DBLP:conf/ijcai/KolbTPR18}, SAT solving~\cite{DBLP:conf/aaai/KumarKTR20}, or MAX-CSP~\cite{DBLP:conf/cp/TsourosS21}. An interesting, perhaps more immediately practical idea has been the extraction of formulas and structure from spreadsheets~\cite{DBLP:conf/aaai/0003KGR21,DBLP:conf/ida/CoetseeKHR21,DBLP:journals/ml/KolbPGR17,DBLP:conf/cikm/0001KGR17}. Another promising approach is limiting the acquisition problem to a specific problem domain, like time tabling~\cite{DBLP:conf/ictai/KumarTCR19} or scheduling~\cite{DBLP:conf/aips/SenderovichBB19}. This allows more specialized methods to be applied, and provides a vocabulary and background knowledge that can be exploited.

There has been relatively little work on acquiring size independent models from multiple problem instances~\cite{DBLP:conf/ictai/LallouetLMV10}, and only more recently work on constraint acquisition with noisy data~\cite{DBLP:journals/amai/PrestwichFOB21,DBLP:conf/ictai/Prestwich21,DBLP:conf/ecai/Prestwich20}, for which a realistic use case has not yet been clearly defined. The case of active acquisition where certain queries cannot be answered by the user is explored in~\cite{DBLP:conf/cp/TsourosSB20}.

The major influences of the current system are the ModelSeeker~\cite{DBLP:conf/cp/BeldiceanuS12,DBLP:conf/cp/BeldiceanuILS13} and the Global Constraint catalog~\cite{catalog2012,DBLP:journals/corr/ArafailovaBDCFR16}. ModelSeeker was acquiring instance specific constraint models expressed as a conjunction of global constraints based on few positive examples of unstructured solutions. We extend its idea of a partition generator to allow generalization of the patterns found, while dealing with structured data for both solutions and input data. Other elements of ModelSeeker, like the constraint subsumption check and ranking of candidate constraints have been reimplemented, but are fundamentally unchanged. The current system does not cover all constraints in the global constraint catalog, restricting itself to those which are already provided in MiniZinc, or for which simple decompositions could be provided, there are in total 50 global constraints defined in the system.

The evaluation problems for the current system are taken from~\cite{simonis_helmut_2021_5155465}, which defined the input format and evaluation strategy followed.



\section{A Running Example}
\label{sec:running example}

To illustrate the approach taken, we use a running example, which is based on the BIBD (Balanced Incomplete Block Design)\cite{colbourn1996crc} problem. A BIBD is defined as an arrangement of $v$ distinct objects into $b$ blocks such that each block contains exactly $k$ distinct objects, each object occurs in exactly $r$ different blocks, and every two distinct objects occur together in exactly $\lambda$ blocks. The values for $b$ and $r$ can be deduced from the values $\lambda$, $v$, and $k$. Instead of this integer formulation, a 0/1 incidence matrix with $v$ rows and $b$ columns is often used as an equivalent representation. The constraints then state that each row contains $r$ ones, each column contains $k$ ones, and any two rows share ones in $\lambda$ columns. This form has been widely used in Constraint Programming~\cite{csplib:prob028,DBLP:journals/constraints/PrestwichHSRT12}, and features as problem 12 in our evaluation set. Figure~\ref{tab:bibd} shows the first example solution of the (2,6,3) BIBD on the left. On the right we see the three corresponding input parameters $(\lambda, v, k)$ in the JSON syntax of~\cite{simonis_helmut_2021_5155465}. Altogether, we are given 20 problem instances of different sizes, each with a set of solutions and non-solutions. While for some problem sizes all solutions are given, for some of the larger problem sizes only a fraction of the complete solution set is provided. The non solutions were generated either by taking random assignment of the variables, or modifying existing solutions in a few places, while checking that they really are non solutions. Note that our acquisition process, given the 0/1 representation of the problem, will acquire the model in terms of 0/1 variables, and not in terms of the integer variables of the original definition. 

\begin{table}
\caption{\label{tab:bibd}Example Problem Instance (Problem 12, BIBD)}
\begin{tabular}{cc}
Positive Example & Input Parameters \\
\begin{tabular}{*{10}{r}}
1&1&1&1&1&0&0&0&0&0\\
1&1&0&0&0&1&1&1&0&0\\
1&0&1&0&0&1&0&0&1&1\\
0&1&0&1&0&0&1&0&1&1\\
0&0&1&0&1&0&1&1&1&0\\
0&0&0&1&1&1&0&1&0&1\\
\end{tabular}
 
&
\begin{lstlisting}[language=json]
    "inputData": {
        "lambda": 2,
        "v": 6,
        "k": 3
    },
\end{lstlisting}
\end{tabular}
\end{table}

We are asked to find a model of the problem that, given new values for the input parameters, finds a solution matrix of 0/1 values of the correct size that satisfies all constraints of the problem.

\section{System Description}
\label{sec:system description}

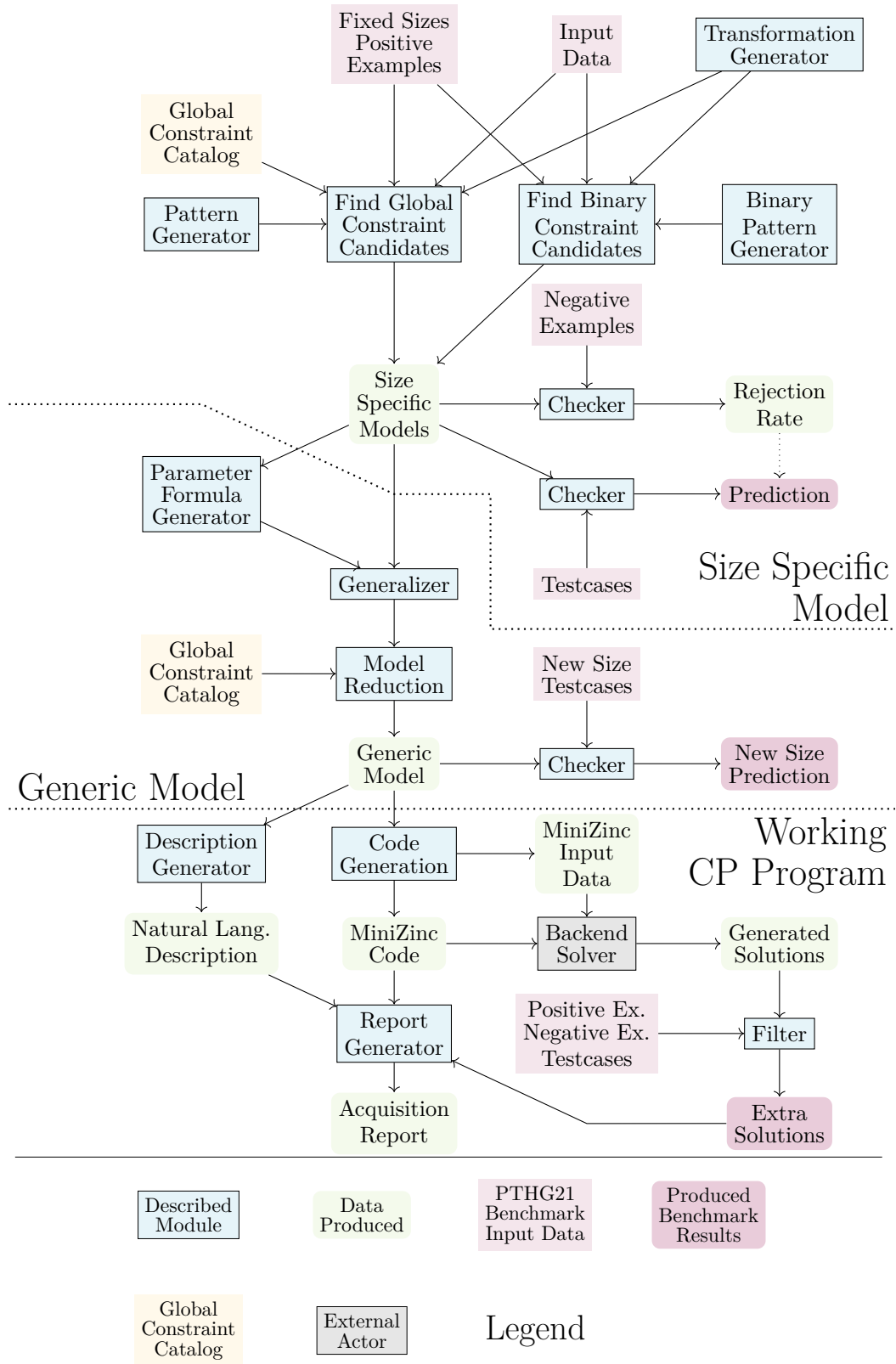
\begin{figure}
\caption{\label{fig:architecture}System Architecture and Evaluation Methodology}
\begin{tikzpicture}[xscale=3,yscale=1.4,
  cat/.style={draw=black,fill=insight-royalblue!10},
  catalog/.style={fill=insight-yellow!10},
  produced/.style={fill=insight-lime!10,rounded corners},
  benchresult/.style={fill=insight-maroon!20,rounded corners},
  external/.style={draw=black,fill=black!10},
  bench/.style={fill=insight-maroon!10}]
  \node[catalog] (catalog) at (1,20) {\shortstack{Global\\Constraint\\Catalog}};
  \node[bench] (posexamples) at (2,21) {\shortstack{Fixed Sizes\\Positive\\Examples}};
  \node[bench] (inputdata) at (3,21) {\shortstack{Input\\Data}};
  \node[cat] (transformation) at (4,21) {\shortstack{Transformation\\Generator}};
  \node[cat] (patterngenerator) at (1,19) {\shortstack{Pattern\\Generator}};
  \node[cat] (seeker) at (2,19) {\shortstack{Find Global\\Constraint\\Candidates}};
  \node[cat] (binaryseeker) at (3,19) {\shortstack{Find Binary\\Constraint\\Candidates}};
  \node[cat] (binarypatterngenerator) at (4,19) {\shortstack{Binary\\Pattern\\Generator}};
  \node[produced] (models) at (2,17) {\shortstack{Size\\Specific\\Models}};
  \node[bench] (negexamples) at (3,18) {\shortstack{Negative\\Examples}};
  \node[cat] (checker) at (3,17) {Checker};
  \node[produced] (rejectionrate) at (4,17) {\shortstack{Rejection\\Rate}};
  \node[cat] (checker2) at (3,16) {Checker};
  \node[bench] (testcases) at (3,15) {Testcases};
  \node[benchresult] (prediction) at (4,16) {Prediction};
    \draw[->] (catalog) -- (seeker);
  \draw[->] (posexamples) -- (seeker);
  \draw[->] (inputdata) -- (seeker);
  \draw[->] (patterngenerator) -- (seeker);
  \draw[->] (posexamples) -- (binaryseeker);
  \draw[->] (inputdata) -- (binaryseeker);
\draw[->] (transformation) -- (binaryseeker);
\draw[->] (transformation) -- (seeker);
  \draw[->] (binarypatterngenerator) -- (binaryseeker);
  \draw[->] (seeker) -- (models);
  \draw[->] (binaryseeker) -- (models);
  \draw[->] (models) -- (checker);
  \draw[->] (negexamples) -- (checker);
  \draw[->] (checker) -- (rejectionrate);
  \draw[->] (models) -- (checker2);
  \draw[->] (testcases) -- (checker2);
  \draw[->] (checker2) -- (prediction);
  \draw[dotted,->] (rejectionrate) -- (prediction);

  \node[cat] (formula) at (1,16) {\shortstack{Parameter\\Formula\\Generator}};
  \node[cat] (generalizer) at (2,15) {Generalizer};
  \node[catalog] (catalog2) at (1,14) {\shortstack{Global\\Constraint\\Catalog}};
  \node[cat] (reduction) at (2,14) {\shortstack{Model\\Reduction}};
  \node[produced] (genericmodel) at (2,13) {\shortstack{Generic\\Model}};
  \node[bench] (newsizetestcases) at (3,14) {\shortstack{New Size\\Testcases}};
  \node[cat] (checker3) at (3,13) {Checker};
  \node[benchresult] (newsizeprediction) at (4,13) {\shortstack{New Size\\Prediction}};
  \draw[->] (formula) -- (generalizer);
  \draw[->] (generalizer) -- (reduction);
  \draw[->] (reduction) -- (genericmodel);
  \draw[->] (catalog2) -- (reduction);
  \draw[->] (genericmodel) -- (checker3);
  \draw[->] (newsizetestcases) -- (checker3);
  \draw[->] (checker3) -- (newsizeprediction);

  \node[cat] (descriptiongenerator) at (1,12) {\shortstack{Description\\Generator}};
  \node[cat] (codegeneration) at (2,12) {\shortstack{Code\\Generation}};
  \node[produced] (generatedinput) at (3,12) {\shortstack{MiniZinc\\Input\\Data}};
  \node[produced] (description) at (1,11) {\shortstack{Natural Lang.\\Description}};
  \node[produced] (code) at (2,11) {\shortstack{MiniZinc\\Code}};
  \node[cat] (reportgen) at (2,10) {\shortstack{Report\\Generator}};
  \node[produced] (report) at (2,9) {\shortstack{Acquisition\\Report}};
  \node[external] (solver) at (3,11) {\shortstack{Backend\\Solver}};
  \node[produced] (solutions) at (4,11) {\shortstack{Generated\\Solutions}};
  \node[cat] (filter) at (4,10) {Filter};
  \node[bench] (benchdata) at (3,10) {\shortstack{Positive Ex.\\Negative Ex.\\Testcases}};
  \node[benchresult] (extra) at (4,9) {\shortstack{Extra\\Solutions}};
   \draw[->] (descriptiongenerator) -- (description);
  \draw[->] (codegeneration) -- (code);
  \draw[->] (codegeneration) -- (generatedinput);
  \draw[->] (code) -- (solver);
  \draw[->] (generatedinput) -- (solver);
  \draw[->] (code) -- (reportgen);
  \draw[->] (description) -- (reportgen);
  \draw[->] (reportgen) -- (report);
  \draw[->] (solver) -- (solutions);
  \draw[->] (solutions) -- (filter);
  \draw[->] (benchdata) -- (filter);
  \draw[->] (filter) -- (extra);
  \draw[->] (extra) -- (3,9) -- (reportgen);

  \draw[->] (models) -- (formula);
  \draw[->] (models) -- (generalizer);

  \draw[->] (genericmodel) -- (codegeneration);
  \draw[->] (genericmodel) -- (descriptiongenerator);
  
  \draw[dotted,thick] (0,17) -- (1,17) -- (2,16) -- (2.5,16) -- (2.5,14.5) -- node[near end,above] {\LARGE \shortstack[r]{Size Specific\\Model}} (4.6,14.5);
  \draw[dotted,thick] (0,12.5) -- node[at start,above right] {\LARGE Generic Model} node[at end,below left] {\LARGE \shortstack[r]{Working\\CP Program}} (4.6,12.5);
\end{tikzpicture}

{\scalebox{0.9}{
\begin{tikzpicture}[xscale=3,yscale=2,
  cat/.style={draw=black,fill=insight-royalblue!10},
  catalog/.style={fill=insight-yellow!10},
  produced/.style={fill=insight-lime!10,rounded corners},
  benchresult/.style={fill=insight-maroon!20,rounded corners},
  external/.style={draw=black,fill=black!10},
  action/.style={fill=insight-burntorange!10},
  bench/.style={fill=insight-maroon!10}]
  \node[cat] () at (1,2) {\shortstack{Described\\Module}};
  \node[produced] () at (2,2) {\shortstack{Data\\Produced}};
  \node[bench] () at (3,2) {\shortstack{PTHG21\\Benchmark\\Input Data}};
  \node[benchresult] () at (4,2) {\shortstack{Produced\\Benchmark\\Results}};
  \node[catalog] () at (1,1) {\shortstack{Global\\Constraint\\Catalog}};
  \node[external] () at (2,1) {\shortstack{External\\Actor}};
\node[] () at (3,1) {\LARGE Legend};
  \draw[black] (0,2.5) -- (5,2.5);
\end{tikzpicture}
}}
\end{figure}

We now start with the description of the constraint acquisition tool. Its overall architecture is presented in Figure~\ref{fig:architecture}, which shows the main elements of the tool and how it is used in the evaluation process. The figure shows that a large number of individual components are needed to run and evaluate the acquisition tool. Most components are relatively simple, separating the overall acquisition process into more manageable pieces.

There are three main parts of the system, from top to bottom. The first part produces instance specific models for each given problem instance, which can be used to check provided test cases for the given problem sizes. 

The second part generalizes these size specific models into one generic model that applies for all sizes. At this point we try to reduce the model by eliminating subsumed constraints, and by ranking constraints according to their interest.

The third module produces an executable model in MiniZinc, while also providing a natural language description and a report on the acquisition process for the user. The generated code can then be used to find new solutions for unseen problem instances.

Figure~\ref{fig:architecture} also shows the data flow in the evaluation for the PTHG21 Challenge. The tool is used both for classifying test cases as solutions or non-solutions, and to produce new, additional solutions that were not given either as samples or test cases.
In the following subsections we discuss each of these high-level modules in more detail. 

\subsection{Generating Size Specific Models}
\label{sec:size specific models}

This part of the acquisition process is quite similar to the ModelSeeker system, with some variations to allow generalization in the second stage. We build a size specific model of the given problem by looking for conjunctions of global constraints that apply to multiple, structured subsets of decision variables of the problem. We assume that these subsets are either defined by the problem structure (e.g. the rows of a matrix of variables), or given as input data (e.g. defining an index set in the input data which selects the elements of a matrix for which a constraint holds). This fundamental simplification makes the acquisition process manageable even for large problem instances. We do not have to search through an exponential number of possible subsets of the variables to identify the scope of a constraint. By considering index sets given as input data, we extend the type of constraints that can be found compared to the ModelSeeker, which had no general concept of input data, and could therefore only recognize regular constraint structures. 

\subsubsection{Transformation Generator}
The way the solution is presented may not be the best way for expressing some of the constraints of the system. In particular some variables of the problem may not be present in the solution description as they are dependent on other variables. We consider the following transformations:
\begin{description}
\item[difference] A relatively common transformation occurs when a constraint is expressed on a set of difference variables $x_{i}-x_{j}$ over a set of given decision variables $x_{i}$ instead of the decision variables themselves. Examples are the Golomb Ruler problem or Costas Array in the benchmark set. A number of variations (absolute value of differences, differences between consecutive, or between all different variables) are possible and must be treated as specific cases.
\item[absolute value] Another transformation is the use of the absolute value of a decision variable instead of the variable itself.
\item[cost variables] A very common transformation occurs when, based on a decision variable, we encounter a cost variable, which depends on the decision value. We assume that the input data provide a cost vector or matrix $d$ that define the cost value for each value of the decision variables. We consider constraints of the form $c_{i} = d_{i}*x_{i}$ as well as $c_{i} = d_{i,x{_i}}$ for implicit cost variables $c_{i}$ which often occur in problems (like Warehouse Location in the benchmark set). 
\item[projection] The most general form of a transformation is a projection, when we use the projected value of a decision variable instead of the variable itself. A typical example would be using the machine type instead of the assigned machine itself to determine some resource constraint in a scheduling problem. As we assume the input data to describe the problem completely, we must find in the input data a vector (or matrix) which transforms a value for a decision variable into a projected domain. This imposes constraints on the size of the matrix that may be used in this way.   
\end{description}

We generate all of these transformed variables "on spec", in case we will use them to define some constraint of the problem, and retain only those transformations that were actually used.

ModelSeeker was using additional transformations to deal with different problem representations, e.g. using 0/1 variables instead of integer decision variables, or having multiple representations of successor variables in graph constraints. These are currently not included in our system, but could be added when required for future problems.

\subsubsection{Pattern Generator}

The concept of pattern (or partition) generator was introduced for ModelSeeker in~\cite{Beldiceanu2011}. Patterns were specified by type and various parameter values, which allowed multiple interpretations of the same unstructured list of variables. This flexibility has a price, it is very difficult to generalize the partitions used for different problem sizes. We here use a simpler approach, which does not produce all partitions used in ModelSeeker, but were the patterns are trivial to generalize, and which exploits the structure of the data and samples given as input.
Instead of formally defining each pattern, we explain the concept on an example. 
\begin{figure}[htbp]
\caption{\label{fig:pattern generator}Pattern Generator for Matrix}
\scalebox{0.55}{
  \begin{tikzpicture}[
  matrix/.style={draw=black!20,rounded corners},
  subset/.style={draw=black!10,fill=insight-royalblue!10},
  subsetsecondary/.style={draw=black!10,fill=insight-lime!10},
  empty/.style={draw=black!10},
  selected/.style={fill=insight-royalblue!20},
  secondary/.style={fill=insight-lime!20},
  dvar/.style={fill=insight-burntorange!20},
  subsetline/.style={draw=insight-royalblue!10,very thick},
  selectedline/.style={draw=insight-royalblue!30,very thick},
    ]
    \node[left] at (-0.3,7.5) {\Large Signature};
    \node[] at (2.5,7.5) {\Large Matrix/Matrices};
    \node[left] at (-0.3,5.5) {\Large \shortstack[r]{List\\List+Dvar}};
    \node[left] at (-0.3,1.5) {\Large \shortstack[r]{List\\List+Dvar}};
    \node[left] at (-0.3,-2.5) {\Large \shortstack[r]{List+List\\List+List+Dvar}};
    \node[left] at (-0.3,-6.5) {\Large \shortstack[r]{List+List\\List+List+Dvar}};
    \node[left] at (-0.3,-10.5) {\Large \shortstack[r]{Matrix\\Matrix+Dvar}};
    \node[left] at (-0.3,-14.5) {\Large \shortstack[r]{List+List\\List+List+Dvar}};
  \begin{scope}[shift={(0,4)}]
  \draw[matrix] (0,0) rectangle (5,3);
  \draw[selected] (0.2,0.2) rectangle (4.8,2.8);
  \node[below] at (2.5,0) {All};
  \end{scope}
  \begin{scope}[shift={(6,4)}]
  \draw[matrix] (0,0) rectangle (5,3);
  \draw[matrix] (5.5,0) rectangle (6,2);
  \draw[selected] (0.2,0.2) rectangle (0.4,0.4);
  \draw[selected] (1.2,0.5) rectangle (1.4,0.7);
  \draw[selected] (1.2,2.2) rectangle (1.4,2.4);
  \draw[selected] (3.7,0.8) rectangle (3.9,1.0);
  \draw[selected] (4.0,2.2) rectangle (4.2,2.4);
  \draw[->] (5.7,0.2) -- (0.4,0.3);
  \draw[->] (5.7,0.4) -- (1.4,0.6);
  \draw[->] (5.7,0.6) -- (3.9,0.9);
  \draw[->] (5.7,0.8) -- (1.4,2.3);
  \draw[->] (5.7,1.6) -- (4.2,2.3);
  \node[below] at (2.5,0) {Indexed};
  \end{scope}
  \begin{scope}[shift={(13,4)}]
  \draw[matrix] (0,0) rectangle (4,3);
  \draw[selected] (0.2,0.2) rectangle (0.4,0.4);
  \draw[selected] (1.2,0.5) rectangle (1.4,0.7);
  \draw[selected] (1.2,2.2) rectangle (1.4,2.4);
  \draw[selected] (3.7,0.8) rectangle (3.9,1.0);
  \draw[selected] (2.0,2.2) rectangle (2.2,2.4);
  \draw[matrix] (4.5,0) rectangle (8.5,3);
  \begin{scope}[shift={(4.5,0)}]
  \draw[selected] (0.2,0.2) rectangle (0.4,0.4);
  \draw[selected] (1.2,0.5) rectangle (1.4,0.7);
  \draw[selected] (1.2,2.2) rectangle (1.4,2.4);
  \draw[selected] (3.7,0.8) rectangle (3.9,1.0);
  \draw[selected] (2.0,2.2) rectangle (2.2,2.4);
  \end{scope}
  \node[below] at (4,0) {Indexed Matrix};
  \end{scope}
  \begin{scope}[shift={(0,0)}]
  \draw[matrix] (0,0) rectangle (5,3);
  \draw[subset] (0.2,0.2) rectangle (4.8,0.4);
  \draw[subset] (0.2,0.5) rectangle (4.8,0.7);
  \draw[selected] (0.2,1.5) rectangle (4.8,1.7);
  \draw[subset] (0.2,2.6) rectangle (4.8,2.8);
  \node[below] at (2.5,0) {AllRows};
  \end{scope}
  \begin{scope}[shift={(6,0)}]
  \draw[matrix] (0,0) rectangle (5,3);
  \draw[subset] (0.2,0.2) rectangle (0.4,2.8);
  \draw[subset] (0.5,0.2) rectangle (0.7,2.8);
  \draw[selected] (1.5,0.2) rectangle (1.7,2.8);
  \draw[subset] (4.6,0.2) rectangle (4.8,2.8);
  \node[below] at (2.5,0) {AllColumns};
  \end{scope}
  \begin{scope}[shift={(12,0)}]
  \draw[matrix] (0,0) rectangle (3,3);
  \draw[subset] (0.2,0.2) rectangle (0.8,0.8);
  \draw[selected] (1.2,1.2) rectangle (1.8,1.8);
  \draw[subset] (1.2,0.2) rectangle (1.8,0.8);
  \draw[subset] (2.2,2.2) rectangle (2.8,2.8);
  \node[below] at (1.5,0) {Blocks};
  \node[above] at (1.5,3) {square};
  \end{scope}
  \begin{scope}[shift={(16,0)}]
  \draw[matrix] (0,0) rectangle (3,3);
  \draw[subsetline] (0.2,0.2) -- (2.8,2.8);
  \draw[selectedline] (0.2,2.8) -- (2.8,0.2);
  \node[below] at (1.5,0) {Diagonals};
  \node[above] at (1.5,3) {square};
  \end{scope}
  \begin{scope}[shift={(0,-4)}]
  \draw[matrix] (0,0) rectangle (5,3);
  \draw[empty] (0.2,0.2) rectangle (4.8,0.4);
  \draw[empty] (0.2,0.5) rectangle (4.8,0.7);
  \draw[selected] (0.2,1.5) rectangle (4.8,1.7);
  \draw[secondary] (0.2,1.8) rectangle (4.8,2.0);
  \draw[empty] (0.2,2.6) rectangle (4.8,2.8);
  \node[below] at (2.5,0) {ConsecutiveRows};
  \draw[|->] (2.5,1.9) -- (2.5,2.7);
  \draw[->] (2.5,1.9) -- (2.5,0.6);
  \end{scope}
  \begin{scope}[shift={(6,-4)}]
  \draw[matrix] (0,0) rectangle (5,3);
  \draw[subset] (0.2,0.2) rectangle (4.8,0.4);
  \draw[subset] (0.2,0.5) rectangle (4.8,0.7);
  \draw[selected] (0.2,1.2) rectangle (4.8,1.4);
  \draw[secondary] (0.2,2.1) rectangle (4.8,2.3);
  \draw[empty] (0.2,2.6) rectangle (4.8,2.8);
  \node[below] at (2.5,0) {OrderedRows};
  \draw[|->] (2.5,2.2) -- (2.5,2.7);
  \draw[->] (2.5,2.2) -- (2.5,0.6);
  \end{scope}
  \begin{scope}[shift={(12,-4)}]
  \draw[matrix] (0,0) rectangle (5,3);
  \draw[subset] (0.2,0.2) rectangle (4.8,0.4);
  \draw[subset] (0.2,0.5) rectangle (4.8,0.7);
  \draw[selected] (0.2,1.2) rectangle (4.8,1.4);
  \draw[secondary] (0.2,2.1) rectangle (4.8,2.3);
  \draw[subset] (0.2,2.6) rectangle (4.8,2.8);
  \draw[|->] (2.5,2.2) -- (2.5,2.7);
  \draw[->] (2.5,2.2) -- (2.5,0.3);
  \node[below] at (2.5,0) {DifferentRows};
  \end{scope}
    \begin{scope}[shift={(0,-8)}]
  \draw[matrix] (0,0) rectangle (5,3);
  \draw[empty] (0.2,0.2) rectangle (0.4,2.8);
  \draw[empty] (0.5,0.2) rectangle (0.7,2.8);
  \draw[selected] (1.5,0.2) rectangle (1.7,2.8);
  \draw[secondary] (1.2,0.2) rectangle (1.4,2.8);
  \draw[empty] (4.3,0.2) rectangle (4.5,2.8);
  \draw[empty] (4.6,0.2) rectangle (4.8,2.8);
  \draw[|->] (1.3,1.5) -- (0.3,1.5);
  \draw[|->] (1.3,1.5) -- (4.4,1.5);
  \node[below] at (2.5,0) {ConsecutiveColumns};
  \end{scope}
    \begin{scope}[shift={(6,-8)}]
  \draw[matrix] (0,0) rectangle (5,3);
  \draw[empty] (0.2,0.2) rectangle (0.4,2.8);
  \draw[empty] (0.5,0.2) rectangle (0.7,2.8);
  \draw[secondary] (1.2,0.2) rectangle (1.4,2.8);
  \draw[selected] (1.8,0.2) rectangle (2.0,2.8);
  \draw[subset] (4.3,0.2) rectangle (4.5,2.8);
  \draw[subset] (4.6,0.2) rectangle (4.8,2.8);
  \draw[|->] (1.3,1.5) -- (0.3,1.5);
  \draw[|->] (1.3,1.5) -- (4.4,1.5);
  \node[below] at (2.5,0) {OrderedColumns};
  \end{scope}
    \begin{scope}[shift={(12,-8)}]
  \draw[matrix] (0,0) rectangle (5,3);
  \draw[subset] (0.2,0.2) rectangle (0.4,2.8);
  \draw[subset] (0.5,0.2) rectangle (0.7,2.8);
  \draw[secondary] (1.2,0.2) rectangle (1.4,2.8);
  \draw[selected] (1.8,0.2) rectangle (2.0,2.8);
  \draw[subset] (4.3,0.2) rectangle (4.5,2.8);
  \draw[subset] (4.6,0.2) rectangle (4.8,2.8);
  \draw[|->] (1.3,1.5) -- (0.3,1.5);
  \draw[|->] (1.3,1.5) -- (4.7,1.5);
  \node[below] at (2.5,0) {DifferentColumns};
  \end{scope}

  \begin{scope}[shift={(0,-12)}]
  \draw[matrix] (0,0) rectangle (5,3);
  \draw[selected] (0.2,0.2) rectangle (4.8,0.4);
  \draw[selected] (0.2,0.5) rectangle (4.8,0.7);
  \draw[selected] (0.2,0.8) rectangle (4.8,1.0);
  \draw[selected] (0.2,1.1) rectangle (4.8,1.3);
  \draw[selected] (0.2,1.8) rectangle (4.8,2.0);
  \draw[selected] (0.2,2.3) rectangle (4.8,2.5);
  \draw[selected] (0.2,2.6) rectangle (4.8,2.8);
  \node[below] at (2.5,0) {Matrix};
  \end{scope}
  \begin{scope}[shift={(6,-12)}]
  \draw[matrix] (0,0) rectangle (5,3);
  \draw[selected] (0.2,0.2) rectangle (0.4,2.8);
  \draw[selected] (0.5,0.2) rectangle (0.7,2.8);
  \draw[selected] (1.5,0.2) rectangle (1.7,2.8);
  \draw[selected] (4.3,0.2) rectangle (4.5,2.8);
  \draw[selected] (4.6,0.2) rectangle (4.8,2.8);
  \node[below] at (2.5,0) {TransposedMatrix};
  \end{scope}
  \begin{scope}[shift={(0,-16)}]
  \draw[matrix] (0,0) rectangle (2.5,3);
  \draw[matrix] (3,0) rectangle (5.5,3);
  \draw[subset] (0.2,0.2) rectangle (2.3,0.4);
  \draw[subset] (0.2,0.5) rectangle (2.3,0.7);
  \draw[selected] (0.2,1.5) rectangle (2.3,1.7);
  \draw[subset] (0.2,2.6) rectangle (2.3,2.8);

   \draw[subsetsecondary] (3.2,0.2) rectangle (5.3,0.4);
  \draw[subsetsecondary] (3.2,0.5) rectangle (5.3,0.7);
  \draw[secondary] (3.2,1.5) rectangle (5.3,1.7);
  \draw[subsetsecondary] (3.2,2.6) rectangle (5.3,2.8);
\node[below] at (2.75,0) {PairedRows};
  \end{scope}
  \begin{scope}[shift={(7,-16)}]
  \draw[matrix] (0,0) rectangle (2.5,3);
  \draw[matrix] (3,0) rectangle (5.5,3);
  
  \draw[subset] (0.2,0.2) rectangle (0.4,2.8);
  \draw[subset] (0.5,0.2) rectangle (0.7,2.8);
  \draw[selected] (1.5,0.2) rectangle (1.7,2.8);
  \draw[subset] (2.1,0.2) rectangle (2.3,2.8);

  \draw[subsetsecondary] (3.2,0.2) rectangle (3.4,2.8);
  \draw[subsetsecondary] (3.5,0.2) rectangle (3.7,2.8);
  \draw[secondary] (4.5,0.2) rectangle (4.7,2.8);
  \draw[subsetsecondary] (5.1,0.2) rectangle (5.3,2.8);

\node[below] at (2.75,0) {PairedColumns};
  \end{scope}


  \end{tikzpicture}
  }
\end{figure}
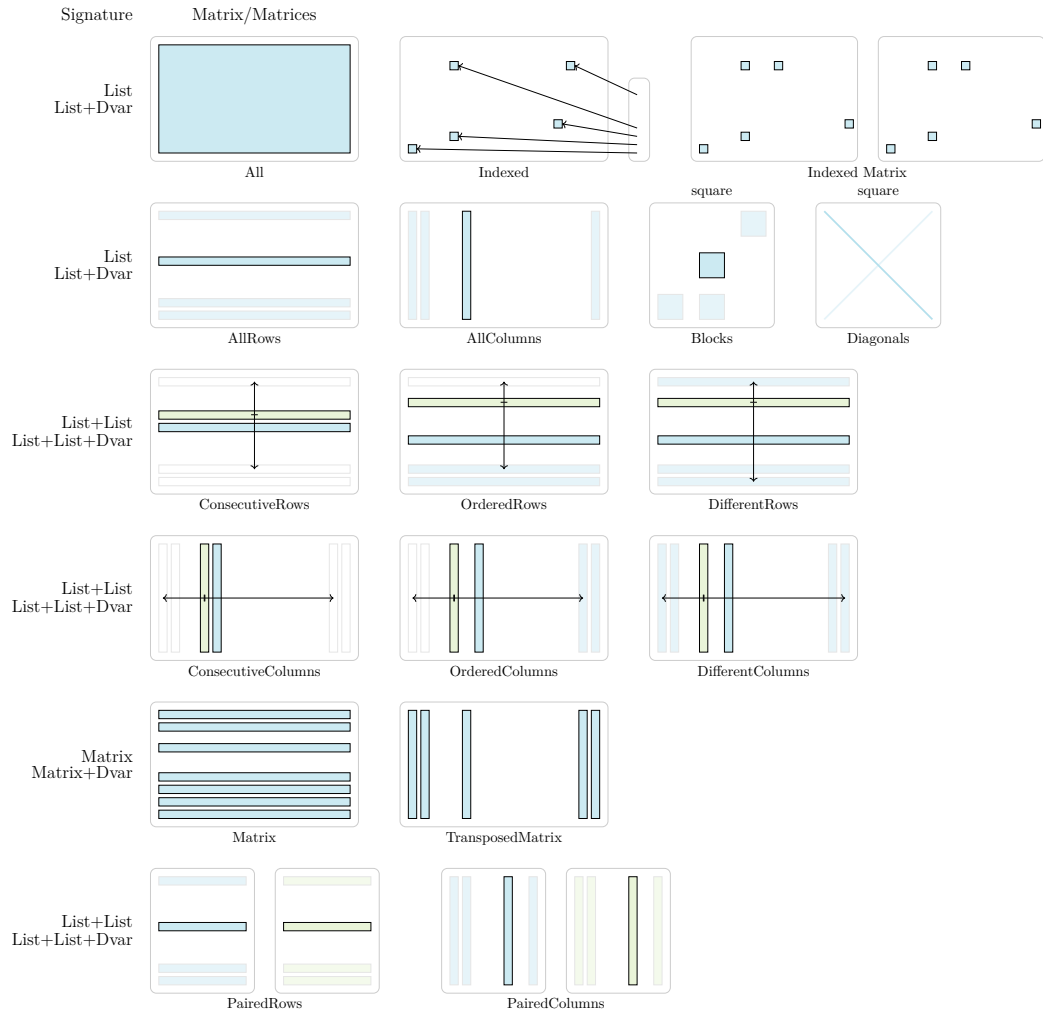

Figure~\ref{fig:pattern generator} shows the patterns that are currently used when dealing with a matrix (or multiple matrices). The first pattern uses the complete matrix and converts it into a single list of all elements. We then have two patterns that use index sets, the first selects the elements of the matrix identified by an index set into a list of variables, the second uses a second matrix as the index set. Elements of the first matrix that share the same index value in the second matrix are grouped together, resulting in a collection of lists of variables, which may or may not have the same size.
The next line shows different ways of grouping elements of a matrix together. The first selects the elements of each row, the second the elements of each column, the thirds builds blocks of equal size, and the last selects the two main diagonals. Note that the last two patterns are only defined for square $n \times n$ matrices, with a block size of $\sqrt{n}$.

All previous patterns could be used for constraints with one list (and optional additional) argument. The next pattern apply to constraints that have two list arguments. We select two rows (resp. columns) of the matrix, either only consecutive rows, or ordered rows, or simply all different rows.

Some constraints use the matrix itself as an argument, we allow that, as well as using the transposed matrix. This can be combined with additional list arguments (not shown).
Finally, we use a pattern where we select corresponding rows or columns of two matrices and use them in constraints with two list arguments. 


\subsubsection{Binary Pattern Generator}

In a similar way, we can define the typical patterns that are used to set up binary constraints. We can of course have constraints between individual variables in a sample, but we assume that constraints over data structures are either of a regular form, or given by some index set in the input data. For a list of variables, we check if the same constraint applies between all different elements of the list, between ordered elements, or consecutive elements only. We also check if corresponding elements of two different lists are constrained by the same binary constraint. 

Some of the binary constraints modelled use a fixed integer constant $c$ in their definition. We assume that only small values of these parameters are relevant, and restrict the range of values tested.

\subsubsection{Finding Candidate Constraint Patterns}
Given some patterns and the original and transformed variables, we then apply each constraint with a matching signature to see if it satisfies each positive sample given. Constraints that do not hold for some positive sample cannot be part of the generated model, as we require consistency of the model with the given samples. When testing the negative samples, we cannot reject the constraints if they do not reject the sample, but we can see if a constraint rejects at least some negative samples. If it does not, it seems too weak to be included in the model.

In our running example we first check which constraints hold over the complete list of all variables. The \texttt{alldifferent} constraint for example does not hold, as the values 0 and 1 are used multiple times. But the constraint $\texttt{nvalue}([x_{1,1},...x_{6,10}],2)$ holds for a count argument of 2. We add the constraint pattern $\langle \texttt{All},x,\texttt{id},\texttt{nvalue}(2)\rangle$ to our candidate constraint pattern set. When testing the rows of the matrix, we again see that the constraint \texttt{alldifferent} does not hold, but the conjunction
\[
\forall_{i \in 1..6}\quad \texttt{sum}([x_{i,1},...,x_{i,10}],5) 
\]
holds with the same value 5 for each row. We therefore remember the constraint candidate $\langle \texttt{EachRow},x,\texttt{id},\texttt{sum}(5)\rangle$ as well. Checking the columns, we find another \texttt{sum} constraint, and checking two rows at the same time, we find that a \texttt{scalar\_product} constraint holds with a value of 2. This constraint works for the patterns \texttt{ConsecutiveRow}, \texttt{OrderedRows} and \texttt{DifferentRows}, so we keep the constraints for each of those patterns. There are many other constraint candidates that are detected, for example the \texttt{notallequal} constraint that holds for each row and column, as well as for the complete set of variables.

The output of this pattern check is a list of all candidate constraint patterns (tuples of pattern, variable, transformations, constraints) that form the model for a given problem instance. Note that at this point we have not yet tried to reduce the set of candidate constraints by subsumption checks or constraint ranking, we also have not yet used the negative examples.

\subsubsection{Checker}
\label{sec:checker}
An important component of the tool is a generalized constraint checker, which applies a combination of pattern, variable, transformation and a set of constraints to a set of samples, and which checks if the constraints are satisfied. This checker relies on an implementation of a checker for each global constraint that is modelled. The Global Constraint Catalog provides such checkers for most constraints, but, as we are using a different implementation language, we had to reimplement these checkers for each constraint.

\subsection{Generalizing Models}
\label{sec:generalizing models}

The next stage of the process is to combine the models created for each problem instance into a generic model, this faces three challenges: 
\begin{itemize}
\item The generic model should only contain constraints that are found in all of the specific models. 
\item In order to be transferable, we have to be able to find all parameter values either directly in the input data, or derive them by some formula from given parameter values.
\item We want to produce a minimal model, and should therefore remove all constraints that are either subsumed by stronger constraints, or that do not contribute to the propagation.
\end{itemize}

These concerns are handled by the following components.

\subsubsection{Generalizer}
\label{sec:generalizer}

By defining a constraint pattern based on structural conditions (e.g. each row of a matrix), instead of numerical parameters, like in ModelSeeker, we can generalize a constraint easily by checking if the same constraint, pattern, and transformation occurs in each instance model. By not performing the subsumption checks already on each instance model, we find all constraints that hold in the generic model without having to consider subsumption rules between them.

In our example, we see that the candidate $\langle \texttt{EachRow},x,\texttt{id},\texttt{sum}(.)\rangle$ is present in each of our instance models, albeit with a different parameter value for the sum. We therefore add this constraint pattern to our generic model. In same same way, we find that a \texttt{sum} constraint over each column, and the \texttt{scalar\_product} constraint are part of the generic model. Constraint patterns that are only in some, but not all, of our instance models are not part of the generic model.

\subsubsection{Formula Generator}
\label{sec:formula generator}

When we generalize the size specific models, we find that different parameter values are used for 
\begin{itemize}
\item the size of lists and matrices
\item the domain bounds of the different sets of variables
\item constraint parameter values
\end{itemize}
To complete our generic model, we have to explain each such parameter in terms of given input data. Figure~\ref{fig:example formula generator} shows an example for our running example, with each line showing on the left the given parameters of the instance, and on the right, the parameter values used in the models. The first line corresponds to the instances of Table~\ref{tab:bibd}. In order to characterize these parameters, we differentiate five cases:
\begin{itemize}
\item If the parameter uses a constant value in every model, then we just use that fixed value. In our example, the domains of the decision variables are always fixed to 0 and 1, shown in blue.
\item If the parameter value used is equal to one of the given input parameters, we use that parameter to explain the value. In Figure~\ref{fig:example formula generator} three parameters (shown in green) use the input parameters directly.
\item As a slight generalization, we consider a set of 20 primitive, one argument functions like $\textrm{npairs}(x) := x(x-1)/2$
, and check if these functions explain the observed values. None of these functions directly explain the missing cases in our example.
\item Finally, we allow nested, complex functions of the form $f_{1} op f_{2}$ with the usual operators $op \in \{+,-,*,/\}$ of increasing size $k$, where each $f_{i}$ is either a primitive function or a complex formula of smaller size. Two such formulas of size 3, shown at the bottom of Figure~\ref{fig:example formula generator}, are needed in our example, one for the number of columns of the matrix, the other to define the row sum value (both in red). 
\item If none of the methods above explains some parameter value, we ask the user for help. The generated program is incomplete, but the user can substitute their own formula to overcome this problem.
\end{itemize}

This brute force, bottom up approach to parameter discovery is very simple, but suffices to find parameter formulas in all given benchmark cases in a few (milli)seconds. More complex equation discovery methods, as for example described in~\cite{DBLP:conf/socs/RazakarisonCBS13}, can be added for more difficult cases in the future.


\begin{figure}[htbp]
\caption{\label{fig:example formula generator}Input Samples for Formula Generator}
\begin{tabular}{lllrrrrrr}\toprule
\multicolumn{3}{c}{Input Parameters} & \multicolumn{3}{c}{Variable Definition} & \multicolumn{3}{c}{Symbolic Values}\\
$\lambda$ & v & k & domain & rows & columns & rowsum & colsum & scalarproduct \\ \midrule
\cellcolor{insight-lime!20}2 & \cellcolor{insight-lime!20}6 & \cellcolor{insight-lime!20}3 & \cellcolor{insight-royalblue!10}0..1 & \cellcolor{insight-lime!20}6 & \cellcolor{insight-burntorange!10}10 & \cellcolor{insight-burntorange!20}5 & \cellcolor{insight-lime!20}3 & \cellcolor{insight-lime!20}2 \\
\cellcolor{insight-lime!20}1 & \cellcolor{insight-lime!20}7 & \cellcolor{insight-lime!20}3 & \cellcolor{insight-royalblue!10}0..1 & \cellcolor{insight-lime!20}7 & \cellcolor{insight-burntorange!10}7 &\cellcolor{insight-burntorange!20} 3& \cellcolor{insight-lime!20}3 & \cellcolor{insight-lime!20}1 \\
\cellcolor{insight-lime!20}4 & \cellcolor{insight-lime!20}6 & \cellcolor{insight-lime!20}3 & \cellcolor{insight-royalblue!10}0..1 & \cellcolor{insight-lime!20}6 & \cellcolor{insight-burntorange!10}20 &\cellcolor{insight-burntorange!20} 10& \cellcolor{insight-lime!20}3& \cellcolor{insight-lime!20}4\\
\cellcolor{insight-lime!20}1 & \cellcolor{insight-lime!20}9 & \cellcolor{insight-lime!20}3 & \cellcolor{insight-royalblue!10}0..1 & \cellcolor{insight-lime!20}9 & \cellcolor{insight-burntorange!10}12 & \cellcolor{insight-burntorange!20}4& \cellcolor{insight-lime!20}3 & \cellcolor{insight-lime!20}1\\
\cellcolor{insight-lime!20}2 & \cellcolor{insight-lime!20}7 & \cellcolor{insight-lime!20}3 & \cellcolor{insight-royalblue!10}0..1 & \cellcolor{insight-lime!20}7 & \cellcolor{insight-burntorange!10}14 & \cellcolor{insight-burntorange!20}6& \cellcolor{insight-lime!20}3& \cellcolor{insight-lime!20}2\\
\cellcolor{insight-lime!20}3 & \cellcolor{insight-lime!20}8 & \cellcolor{insight-lime!20}4 & \cellcolor{insight-royalblue!10}0..1 & \cellcolor{insight-lime!20}8 & \cellcolor{insight-burntorange!10}14 &\cellcolor{insight-burntorange!20} 7& \cellcolor{insight-lime!20}4& \cellcolor{insight-lime!20}3\\
\bottomrule
\end{tabular}
\[
\textrm{columns} = \frac{\lambda * \textrm{npairs}(v)}{\textrm{npairs}(k)} = \frac{\lambda*v(v-1)}{k(k-1)}\quad \textrm{rowsum} = \frac{\lambda(v-1)}{k-1}
\]

\end{figure}

\subsubsection{Model Reduction}
\label{sec:model reduction}

Once we have identified all possible candidate constraint patterns for the generic model, we can perform a model reduction to eliminate unneeded constraints. This reuses the methodology developed for ModelSeeker, using subsumption, contractible and extensible information in the catalog linking different constraints. We for example remove the \texttt{nvalue(2)} constraint over all variables, as it is implied by the domain $0..1$. We also only keep the \texttt{scalar\_product} constraint for the \texttt{AllDifferenentRows} pattern, as the constraints for the patterns \texttt{AllConsecutiveRows} and \texttt{AllOrderedRows} are subsumed.

In addition, we perform a ranking of the constraints to eliminate unlikely model elements. For example, the \texttt{notallequal} constraint over the complete variable set holds in the majority of models, but it is rarely of interest, as it rejects only a minuscule number of possible assignments. On the other hand, we should not reject the constraint altogether. If no other promising constraints are found, and the constraint applies to a smaller subset of the variables, it can be useful in a model. In the benchmark set, one such situation occurs for problem type 07, Schur's Lemma.

\begin{figure}[htbp]
\caption{\label{fig:bibdgraph}Graph Representation of Generic Model}
\begin{tikzpicture}[xscale=3.5,yscale=1.5,
matrix/.style={draw=black!20,fill=insight-royalblue!20,rounded corners},
pattern/.style={draw=black!20,fill=insight-maroon!20},
constraint/.style={draw=black!20,fill=insight-yellow!20},
globalformula/.style={draw=black!20,fill=insight-lime!50},
parameter/.style={draw=black!20,fill=insight-lime!20}]
\node[matrix] (x) at (1,3) {x[1..v,1..cf1]::0..1};
\node[pattern] (rows) at (2,5) {AllRows}; 
\node[pattern] (cols) at (2,4) {AllColumns}; 
\node[pattern] (differentRows) at (2,3) {DifferentRows}; 
\node[pattern] (matrix) at (2,2) {Matrix}; 
\node[pattern] (matrixtransposed) at (2,1) {MatrixTransposed}; 
\node[constraint] (sum1) at (3,5) {sum}; 
\node[constraint] (sum2) at (3,4) {sum}; 
\node[constraint] (scalar) at (3,3) {scalar\_product}; 
\node[constraint] (lexgreater) at (3,2) {lex\_greater}; 
\node[constraint] (lexgreatereq) at (3,1) {lex\_geq}; 
\node[globalformula] (cf3) at (4,5) {$cf3:=\lambda*(v-1)/(k-1)$}; 
\node[parameter] (k) at (4,4) {$k$}; 
\node[parameter] (lambda) at (4,3) {$\lambda$}; 
\node[globalformula] (cf1) at (4,1) {$cf1:=\lambda*p(v)/p(k)$}; 

\draw[->] (x) -- (rows);
\draw[->] (x) -- (cols);
\draw[->] (x) -- (differentRows);
\draw[->] (x) -- (matrix);
\draw[->] (x) -- (matrixtransposed);
\draw[->] (rows) -- (sum1);
\draw[->] (cols) -- (sum2);
\draw[->] (differentRows) to[bend left](scalar);
\draw[->] (differentRows) to[bend right](scalar);
\draw[->] (matrix) -- (lexgreater);
\draw[->] (matrixtransposed) -- (lexgreatereq);
\draw[->] (cf3) -- (sum1);
\draw[->] (k) -- (sum2);
\draw[->] (lambda) -- (scalar);
\end{tikzpicture}
\end{figure}
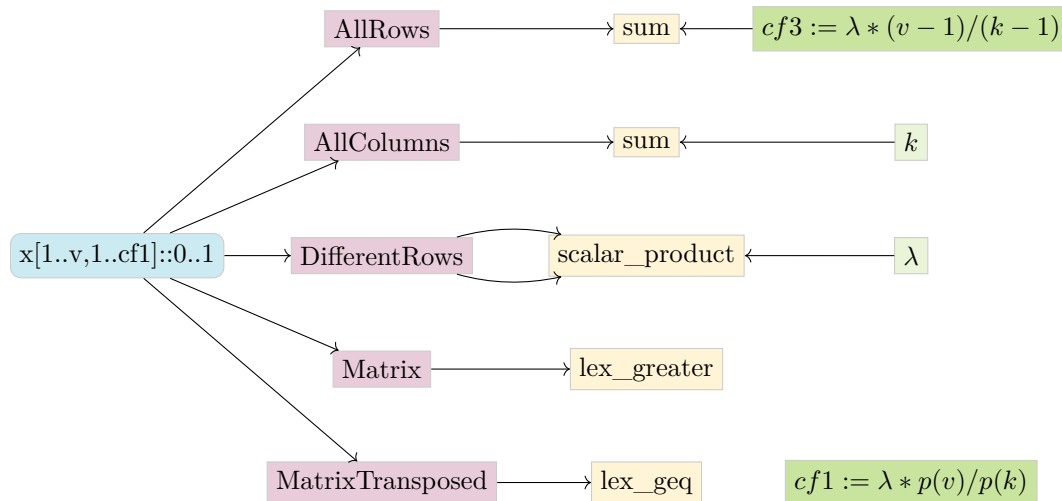

Figure~\ref{fig:bibdgraph} shows a graphical representation of the generic model found for our running example. The model is based on a single matrix (in blue) of 0/1 variables with $v$ rows and $cf1$ columns, we don't use any transformation in this model, and five types of constraints are recognized. Each constraint (in yellow) is represented as a box with a number of ports representing its arguments. There is a \texttt{sum} constraint that applies for each row of the matrix (pattern shown in magenta), with the total given by formula (formulas shown in green) $cf3$, another \texttt{sum} over all columns equal to $k$, the \texttt{scalar\_product} of different rows which is equal to $\lambda$, and two symmetry breaking constraints for the problem, that were found since the samples given did satisfy these constraints. If the problem generator did not use those constraints, we would not identify them for the given problem. This graphical representation is a compact way to show the complete model, and could be used as well by the user to add or remove constraints to a model in a system independent way. A formula $cf2$ was introduced to explain the total number of ones in the matrix, but the corresponding constraint was eliminated in the model reduction.  

\subsection{Producing Working CP Program}
\label{sec:producing program}

The final block of the system is concerned with generating a working constraint program from the generic model, and providing enough information to the user to explain the model and its constraints.

\subsubsection{Code Generator}
\label{sec:code generator}

This major element is code generation, which produces MniZinc code from the generic model. As the MiniZinc version used was not able to read the JSON input data directly, an additional small module was required to transform the given input data in JSON format into an equivalent representation as .dzn files. 
\begin{figure}
\caption{\label{fig:generated code}Generated MiniZinc Code for Example}
\input{bibdcode}
\end{figure}

The code generation itself relies on a library file, which provides implementations of some of the global constraints that are not currently available in MiniZinc directly, and which also provides the primitive functions used in Section~\ref{sec:formula generator} to define new parameters of the system. The code generation is then quite straightforward, first defining the input data used, any computed parameter values, then the definition of the variables used in the system, followed by the constraints that have been selected. Finally, a default search for a feasible solution is specified. If an objective function has been detected, a search for an optimal solution can the given instead. Note that we use generic names for the variables, to avoid name-clashes with reserved words in MiniZinc.

Figure~\ref{fig:generated code} shows the generated code for the running example. For this specific example, we find the three input parameters (v, k, lambda), three dependent parameters (cf1, cf2, cf3), a matrix of 0/1 decision variables $x$, and the selected constraints of the system. Each selection pattern (\texttt{EachRow}, etc) corresponds to a selection mechanism that identifies which variables are in scope of the constraints.

\subsubsection{Description Generator}
\label{sec:description generator}

The second output of the generator phase is a textual description of the constraint model, as given for our running example in Figure~\ref{fig:generated description}. While the inverse problem of extracting a constraint model from an arbitrary textual description is difficult~\cite{DBLP:conf/ijcai/Kiziltan0T16}, we can generate a basic description with a simple, template-based format. Note that more detailed natural language descriptions of the global constraints should be added, we currently only handle some specific constraints with individual templates.  An interesting direction for future research would be to adapt the description to specific problem domains, e.g. describe a model as a scheduling problem if we find the relevant structure.  

\begin{figure}[htbp]
\caption{\label{fig:generated description}Generated Description}
Given integer parameters \emph{k}, \emph{lambda}, and \emph{v}, and introducing symbols
\begin{description}
\item[cf1]  = lambda*npairs(v) div npairs(k)
\item[cf3]  = lambda*nminus1(v) div nminus1(k)
\end{description}
find an assignment for a matrix \emph{x} with \emph{v} rows and \emph{cf1} columns, where each element ranges between \emph{0} and \emph{1},

\noindent such that
\begin{description}
\item[1] the constraint \emph{lexchaingreater(x)} holds for the matrix \emph{x}.

\item[4] the constraint \emph{lexchaingeq(transpose(x))} holds for the transposed matrix \emph{x}.

\item[5] the sum of each row of \emph{x} is equal to \emph{cf3}.

\item[6] the sum of each column of \emph{x} is equal to \emph{k}.

\item[7] the scalar product of every pair of different rows of \emph{x} is equal to \emph{lambda}.

\end{description}

\end{figure}

\subsubsection{Report Generator}
\label{sec: report generator}

In some cases, a user may be satisfied with only the generated code and description, but a lot more information about the acquisition process is available. We therefore generate a report that documents the different steps of the process, and that shows which information was extracted, and which choices were made. It lists the constraints found, showing which constraints were rejected either by subsumption, or low ranking, and which formulas were generated from the given data.

%

\section{Evaluation}
\label{sec:evaluation}

We have evaluated our system on the benchmark problems defined for the PTHG21 Challenge~\cite{simonis_helmut_2021_5155465}. This dataset provides 15 problem, for which multiple instances are considered. For each instance, solutions and non-solutions are provided. The number of solutions given varies widely, some problems only have a single solution, for others only a fraction of all possible solutions are given. 
In addition to the given samples, the data also provide test cases of known and unknown sizes which the system needs to classify as solutions or non-solutions. 

Finally, the participants can enter new solutions found, which are not covered by the given samples or tests, to show that their system can generate previously unseen solutions. The overall test score is a combination of the number of tests labelled correctly, and the number and correctness of any extra solutions provided.

Table~\ref{tab:pthg21problems} lists the problems in the benchmark set, a large number are taken from CSPlib, and some variations of well-known puzzles are included. Different problems test specific aspects of the acquisition challenge, like structured solutions, input data, or implicit variables not presented in the given samples.

\begin{table}[htbp]
\caption{\label{tab:pthg21problems}PTHG21 Problems}
{\scriptsize
\begin{tabular}{*{4}{r}} \toprule
Type & Problem & Source & Features \\ \midrule
1 & Graph Coloring & ALICE, CHIP & graph as data, optimization\\
2 & N-Queens & CSPlib 054& \\
3 & Warehouse Location & CSPlib 034& cost matrix/vector as data, implicit cost variables\\
4 & Golomb Ruler & CSPlib 006& implicit decision variables, optimization\\
5 & Sudoku  & Pre-assignment& pre-assignment as data, single solution\\
6 & Sudoku & No pre-assignment& many solutions\\
7 & Schur's Lemma & CSPlib 015 & non-standard variable pattern, ternary constraint\\
8 & All Interval & CSPlib 007& auxiliary variables\\
10 & Magic Squares & CSPlib 019& implicit formula\\
11 & Orthogonal Latin Squares & Euler & constraint on tuples\\
12 & BIBD & CSPlib 028& 3 parameters, implicit formulas, symmetry breaking\\
13 & Costas Array & CSPlib 076& auxiliary variables, constraint on tuples\\
14 & N-Queens variant& fairy chess piece & non-traditional attack\\
15 & N-Queens variant& \\
16 & N-Queens variant& \\
\bottomrule
\end{tabular}

}
\end{table}

Table~\ref{tab:results} shows the results of testing our system with the dataset. The generic models that we generated are accepting all positive examples, and rejecting all negative examples, showing that the constraints generated are flexible enough to deal with the models encountered. When testing the system against the classification data, we achieve 100\% accuracy, indicating that the models generalize the problem correctly. This does not mean that the generated code matches the original problems perfectly. When we try generate extra solution from the models we encounter two problems:
\begin{enumerate}
\item For the Golomb ruler problem the upper bound of the domain is the value of the optimal solution. We are not able to find a generic formula that predicts the value of the optimal solution for a new problem size (this is unlikely to exist in a simple closed form), so that we cannot generate extra solutions for unknown problem sizes. We might call this an issue with the given problem statement.
\item For problem 16, we find most, but not all constraints of an extended n-queen type puzzle, where the pieces can also attack each other in repeated knight's moves. This constraint cannot be expressed by the current patterns that we consider. When generating new solutions, we miss this constraint, and most of the generated solutions are rejected. By chance, the constraints are not required to decide the test cases, so our model still labels all tests correctly. This highlights that our method (like all passive acquisition methods) heavily depends on the quality of the sample solutions and non solutions to identify the intended model.
\end{enumerate}
For two of the benchmarks (Sudoku and Schur's Lemma), there are no extra solutions that can be found, as the number of solutions is limited. The total time to acquire and generate all benchmark problems was $10$ minutes, using a single core on a laptop. This includes the time required to classify the test cases. Each run of the generated model for a given instance to generate extra solutions was limited to 300 seconds.

\begin{table}[htbp]
\caption{\label{tab:results}Results on PTHG21 Benchmarks}
{\scriptsize
\begin{tabular}{llrrrrrl}\toprule
        &      &           & \multicolumn{3}{c}{Classified Correctly} & &\\
Problem & Description & Instances & \shortstack{Positive\\Samples} & \shortstack{Negative\\Samples} & Tests & \shortstack{Extra\\Solutions} & Note\\ \midrule
1 & Graph Coloring & 10& \cellcolor{insight-lime!20} 100.00 &\cellcolor{insight-lime!20} 100.00 &\cellcolor{insight-lime!10} 100.00 & \cellcolor{insight-lime!10}\\
2 & N-Queens & 10& \cellcolor{insight-lime!20} 100.00 &\cellcolor{insight-lime!20} 100.00 &\cellcolor{insight-lime!10} 100.00 & \cellcolor{insight-lime!10}\\
3 & Warehouse Location & 10& \cellcolor{insight-lime!20} 100.00 &\cellcolor{insight-lime!20} 100.00 &\cellcolor{insight-lime!10} 100.00 & \cellcolor{insight-lime!10}\\
4 & Golomb Ruler & 8& \cellcolor{insight-lime!20} 100.00 &\cellcolor{insight-lime!20} 100.00 &\cellcolor{insight-lime!10} 100.00 & \cellcolor{insight-burntorange} & 1\\
5 & Sudoku (with Hints)& 10& \cellcolor{insight-lime!20} 100.00 &\cellcolor{insight-lime!20} 100.00 &\cellcolor{insight-lime!10} 100.00 & \cellcolor{gray!10}\\
6 & Sudoku (without Hints)& 4& \cellcolor{insight-lime!20} 100.00 &\cellcolor{insight-lime!20} 100.00 &\cellcolor{insight-lime!10} 100.00 & \cellcolor{insight-lime!10}\\
7 & Schur's Lemma & 13& \cellcolor{insight-lime!20} 100.00 &\cellcolor{insight-lime!20} 100.00 &\cellcolor{insight-lime!10} 100.00 & \cellcolor{gray!10}\\
8 & All Interval & 10& \cellcolor{insight-lime!20} 100.00 &\cellcolor{insight-lime!20} 100.00 &\cellcolor{insight-lime!10} 100.00 & \cellcolor{insight-lime!10}\\
10 & Magic Squares & 6& \cellcolor{insight-lime!20} 100.00 &\cellcolor{insight-lime!20} 100.00 &\cellcolor{insight-lime!10} 100.00 & \cellcolor{insight-lime!10}\\
11 & Orthogonal Latin Squares & 2& \cellcolor{insight-lime!20} 100.00 &\cellcolor{insight-lime!20} 100.00 &\cellcolor{insight-lime!10} 100.00 & \cellcolor{insight-lime!10}\\
12 & BIBD & 20& \cellcolor{insight-lime!20} 100.00 &\cellcolor{insight-lime!20} 100.00 &\cellcolor{insight-lime!10} 100.00 & \cellcolor{insight-lime!10}\\
13 & Costas Array & 11& \cellcolor{insight-lime!20} 100.00 &\cellcolor{insight-lime!20} 100.00 &\cellcolor{insight-lime!10} 100.00 & \cellcolor{insight-lime!10}\\
14 & N-Queens Variant & 11& \cellcolor{insight-lime!20} 100.00 &\cellcolor{insight-lime!20} 100.00 &\cellcolor{insight-lime!10} 100.00 & \cellcolor{insight-lime!10}\\
15 & N-Queens Variant & 7& \cellcolor{insight-lime!20} 100.00 &\cellcolor{insight-lime!20} 100.00 &\cellcolor{insight-lime!10} 100.00 & \cellcolor{insight-lime!10}\\
16 & N-Queens Variant & 13& \cellcolor{insight-lime!20} 100.00 &\cellcolor{insight-lime!20} 100.00 &\cellcolor{insight-lime!10} 100.00 & \cellcolor{insight-burntorange}& 2 \\
\bottomrule
\end{tabular}
}
\end{table}

We note that the generated programs for all problems can be run with at least one of the MiniZinc back-end solvers, producing initial solutions in seconds. For some problems, not all solutions could be generated within a 300 second timeout, either because there are too many solutions to enumerate, or the solver backtracks deeply after finding initial solutions. 

The generated code is in most, but not all case comparable to a handwritten model of the problem based on the same decision variables. For the Costas Array problem (problem 13) the generated model using a more elegant \texttt{lexdifferent} constraint on difference tuples, while the handwritten models we have seen decompose this constraint into multiple \texttt{alldifferent} constraints.

\section{Conclusions and Future Work}
\label{sec:conclusion}

While the results of the system are very encouraging, there are a number of fundamental problems that need to be addressed before it can become an operational tool for acquiring constraints model for typical applications.
\begin{description}
\item[optimization problems] With our current approach we can identify typical objective functions, but at this point we do not handle cases where only optimal (or near optimal) solutions are accepted. 
\item[index sets] The current support for index sets is limited. We should add more complicated index filtering which depends on input data, for example selecting all tasks whose duration exceeds a parameter value.
\item[large input datasets] In many practical problems, we are not just given the required input data, but many other items and tables that are not relevant to the solution process. Identifying which parts of the input data are needed to define the model is a major research challenge.
\item[unknown constraints] The system relies on the global constraint catalog to provide a wide variety of global constraints that can be used to model the problem. If we are given a new problem that requires some new global constraint not currently described in the catalog and modelled in our system, we will only be able to partially model the problem. 
\item[user interaction] Our current evaluation focusses on an automated acquisition process, where the role of the user is quite limited. We should investigate how the user can interact with a system to help with the acquisition process, providing useful information while avoiding asking too many or too repetitive questions. 
\end{description}

Overall, the system described shows the feasibility of dealing with a more realistic use case for constraint acquisition, indicating that more powerful modelling tools for dealing with real world problem types and sizes are possible. 

\clearpage
\bibliography{bib}
\end{document}